\documentclass[11pt]{article}

\usepackage[preprint]{acl}

\usepackage{times}
\usepackage{latexsym}
\usepackage[T1]{fontenc}
\usepackage[utf8]{inputenc}
\usepackage{microtype}
\usepackage{inconsolata}
\usepackage{graphicx}
\usepackage{makecell}
\usepackage{tikz}
\usepackage{amsmath,amssymb}
\usepackage{booktabs}
\usepackage{xcolor}
\usepackage{algorithm}
\usepackage{algpseudocode}
\usepackage{subcaption}
\usepackage{placeins}
\usepackage{enumitem}

\newcommand{\Algnameabbr}{\texttt{TACO}}
\newcommand{\Algname}{\textbf{T}ail-\textbf{A}ware \textbf{C}redit calibrati\textbf{O}n}
\hypersetup{pdfauthor={},pdfsubject={},pdfkeywords={}}

\makeatletter
\g@addto@macro\normalsize{%
  \setlength\abovedisplayskip{5pt plus 2pt minus 2pt}%
  \setlength\belowdisplayskip{5pt plus 2pt minus 2pt}%
  \setlength\abovedisplayshortskip{2pt plus 1pt}%
  \setlength\belowdisplayshortskip{3pt plus 1pt minus 1pt}}
\makeatother
\setlist[itemize]{topsep=2pt,partopsep=0pt}

\title{Right Answer, Wrong Tokens: Tail-Aware Credit Calibration for RLVR}
\author{
 \textbf{Xiuyi Lou\textsuperscript{1}\thanks{Equal contribution, ordered by rolling dices.}},
 \textbf{Zicheng Xu\textsuperscript{1}\footnotemark[1]},
 \textbf{Yu-Neng Chuang\textsuperscript{2}},
 \textbf{Hoang Anh Duy Le\textsuperscript{2}},
 \\
 \textbf{Zhaozhuo Xu\textsuperscript{3}},
 \textbf{Guanchu Wang\textsuperscript{4}},
 \textbf{Vladimir Braverman\textsuperscript{1}\thanks{Correspondence to: Vladimir Braverman \href{mailto:vova@cs.jhu.edu}{vova@cs.jhu} \\ \href{mailto:vova@cs.jhu.edu}{.edu}.}}
\\
\\
 \textsuperscript{1}Johns Hopkins University,
 \textsuperscript{2}Rice University,
 \\
 \textsuperscript{3}Workato,
 \textsuperscript{4}University of North Carolina at Charlotte
}

\begin{document}

\maketitle

\begin{abstract}


Although reinforcement learning (RL) improves the reasoning capabilities of large language models (LLMs), GRPO-style methods can suffer from a \textbf{\textit{positive-credit contamination}} failure mode. Specifically, uniform credit assignment can reinforce locally erroneous tail tokens alongside useful tokens within successful trajectories. To address this problem, we propose \Algname{} (\Algnameabbr{}), which calibrates the positive credit assigned to such tokens without token-level error labels. To this end, \Algnameabbr{} lets the policy serve as its own
reference, assessing each sampled token's rarity relative
to the policy's local expectations to compute a tail-risk score. Using this score as a proxy for potential local errors,
\Algnameabbr{} softly reduces positive credit for high tail-risk tokens
while retaining nonzero weights, limiting their reinforcement
without eliminating opportunities to learn useful rare patterns. Experiments across four LLMs and eight reasoning benchmarks
show that \Algnameabbr{} consistently improves over GRPO-style baselines. Notably, \Algnameabbr{} also improves training stability, supporting sustained performance gains in long-horizon RL. The source code is available at \url{https://github.com/xiuyilou/TACO}.
\end{abstract}

\section{Introduction}
Reinforcement learning with verifiable rewards (RLVR) has become a common post-training paradigm for reasoning-oriented large language models (LLMs).
Recent models such as OpenAI's o-series and DeepSeek-R1 demonstrate that large-scale RL post-training can substantially improve performance on automatically verifiable tasks~\citep{openai2024o1system,deepseekai2025r1}.
Early RLVR methods often build on Proximal Policy Optimization (PPO), which typically relies on a learned critic for advantage estimation~\citep{schulman2017ppo,ouyang2022training}.
In contrast, Group Relative Policy Optimization (GRPO) replaces the learned critic with group-relative advantages, reducing value-modeling overhead while achieving strong empirical performance~\citep{shao2024deepseekmath}. The simple yet effective design has made GRPO a widely adopted backbone for reasoning RL.

\begin{figure}[t]
    \centering
    \includegraphics[width=0.92\columnwidth]{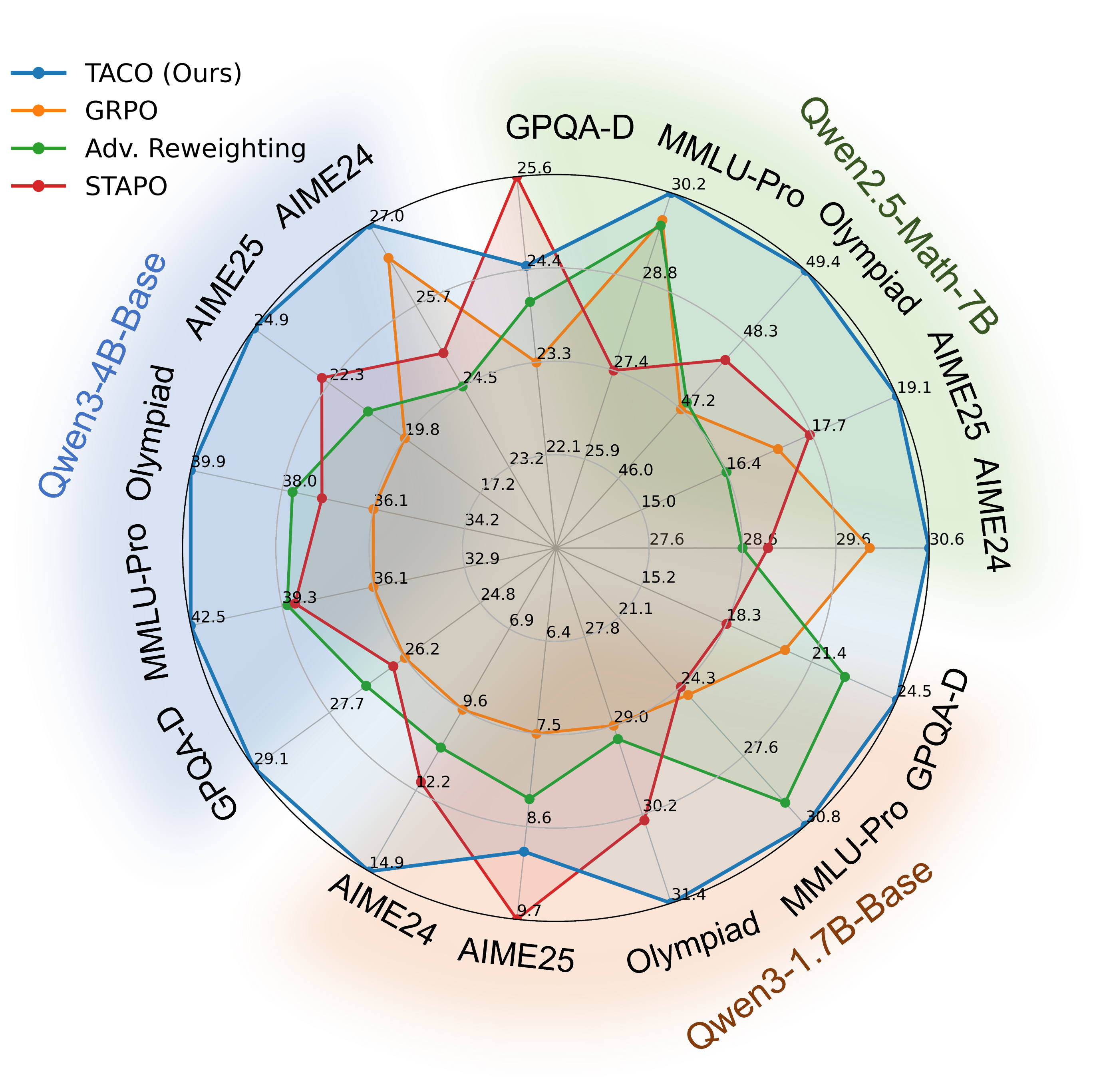}
    \caption{
        \Algnameabbr{} consistently improves over GRPO-style baselines across representative benchmarks and models.
    }
    \label{fig:1}
\end{figure}
However, this simplification computes a single completion-level advantage and broadcasts it to every generated token, even though not every token in a rewarded completion is equally reliable. Specifically, tokens from the low-probability tail of the policy distribution, which are unlikely under the generation context, can be sampled in rewarded completions even when they are semantically irrelevant or erroneous; we refer to such locally unreliable tokens as implausible tail tokens. Yet, these tokens' local effects may be bypassed or overlooked by the overall correct reasoning process and final answer~\citep{uesato2022solving,lightman2023verify}.
These tokens then receive the same positive credit as reliable ones, resulting in indiscriminate reinforcement of
both well-formed reasoning behavior and locally implausible continuations. This exposes locally erroneous continuations to repeated
positive credit during training. We refer to this failure mode as \textbf{\textit{positive-credit contamination}}.

To improve upon the uniform credit assignment in GRPO-style methods,
existing work generally seeks to differentiate token-level updates
based on their estimated importance to the final outcome. One line of
work leverages external signals such as counterfactual analysis,
temporal-difference propagation, or execution feedback to evaluate
each token's contribution and scale its credit
accordingly~\citep{oar2026,egca2026}.
Another line incorporates intrinsic signals such as token entropy or
distributional divergence as proxies for per-token significance,
concentrating updates on tokens deemed critical to the reasoning
process and attenuating routine
ones~\citep{erpo2026,ucas2025}. However, existing methods exhibit two fundamental limitations. First, most of these methods depend on additional inference or auxiliary models, introducing non-trivial resource overhead. Second, existing methods weight tokens by position importance rather than by their contextual reliability~\citep{pmlr-v267-lin25j}. Implausible tail tokens may appear in high-contribution or high-entropy positions that existing methods regard as informative, causing unreliable credit to be mistakenly amplified. Together, these limitations restrict the effectiveness of existing credit-assignment methods in mitigating positive-credit contamination.

To overcome these limitations, we introduce
\Algname{} (\Algnameabbr{}).
\Algnameabbr{} assesses each token relative to the policy's
local expectations, letting the policy serve as its own
reference for positive-credit calibration.
Specifically, \Algnameabbr{} compares each sampled token's
log probability with the local expectation under
the policy's current distribution to compute a tail-risk score.
A token with high tail risk is much less likely than
the policy expects in that context, providing a signal
of a potential local error. \Algnameabbr{} then applies a partial reduction
to the positive credit of high-risk tokens, leaving credit for low-risk tokens and all negative
credit unchanged.
As a result, \Algnameabbr{} limits the reinforcement
of potentially unreliable behaviors while preserving opportunities for useful exploration at negligible additional computational cost.

Empirically, we evaluate \Algnameabbr{} across four LLMs and six mathematical reasoning benchmarks, as well as two out-of-distribution (OOD) scientific reasoning benchmarks.
\Algnameabbr{} consistently improves over GRPO-style baselines, with Figure~\ref{fig:1} highlighting the gains on five representative benchmarks. Our key contributions are:
\begin{itemize}[leftmargin=10pt]
    \itemsep=-1pt
    \item \textbf{Positive-Credit Contamination.} We identify a failure mode of RLVR in which locally implausible tokens receive positive reinforcement.
    \item \textbf{Tail-Aware Credit Calibration.} We introduce \Algnameabbr{}, a context-aware method that calibrates positive token-level credit with negligible computational overhead.
    \item \textbf{Comprehensive Evaluation.} \Algnameabbr{} consistently improves over baselines across multiple benchmarks,
    maintaining stable performance under long-horizon training.
\end{itemize}

\section{Preliminary}

\subsection{Notations.} Let $\pi_\theta$ be an autoregressive policy over vocabulary $\mathcal{V}$.
For a prompt $q$ and completion $o_i = (o_{i,1}, \dots, o_{i,T_i})$,
we denote $c_{i,t} = (q, o_{i,<t})$ as the context at step $t$, and
$p_{i,t}=\pi_\theta(o_{i,t}\mid c_{i,t})$ as the sampled-token probability.
\subsection{Group Relative Policy Optimization}

GRPO is a widely adopted RLVR framework for reasoning-oriented LLM post-training. Given a prompt $q$, GRPO samples a group of $G$ completions $\{o_i\}_{i=1}^{G}$ using the policy $\pi_{\theta_{\mathrm{old}}}$. Each completion $o_i$ receives an outcome-level verified reward $R_i$, and its sequence-level advantage is computed as $\hat{A}_i=\frac{R_i-\mu}{\sigma}$, where $\mu$ and $\sigma$ are the mean and standard deviation of rewards within the sampled group. GRPO then updates the policy by optimizing the following objective,
which aggregates over all generated tokens:
\begin{equation}
\mathcal{J}_{\mathrm{GRPO}}(\theta)=\mathbb{E}_{q,\{o_i\}\sim\pi_{\theta_{\mathrm{old}}}}\Bigl[\frac{1}{T}\sum_{i=1}^{G}\sum_{t=1}^{T_i}\ell_{i,t}\Bigr],
\label{eq:grpo}
\end{equation}
where $q\sim\mathcal{D}$, $\{o_i\}_{i=1}^{G}$ are $G$ responses sampled from $\pi_{\theta_{\mathrm{old}}}(\cdot\mid q)$ with lengths $T_i$, $T=\sum_i T_i$, and $\ell_{i,t}=\ell_{i,t}(\theta;\hat{A}_{i,t})$ denotes the PPO-style clipped surrogate\footnote{The full GRPO loss function can be referred to Equation (1) in the DeepSeek-R1 technical report~\cite{deepseekai2025r1}.}.  Since updates are token-level while advantages are completion-level, GRPO broadcasts the same advantage to every token position: $\hat{A}_{i,t}=\hat{A}_i$. This broadcast rule assigns identical credit regardless of token-level differences. In reasoning traces, this can dilute useful learning signals and lead to suboptimal policy
updates~\citep{int2026intervention}. We examine this issue in detail
below.

\subsection{Positive-Credit Contamination}
\label{sec:pcc}
\begin{figure}[t]
    \centering
    \includegraphics[width=\columnwidth]{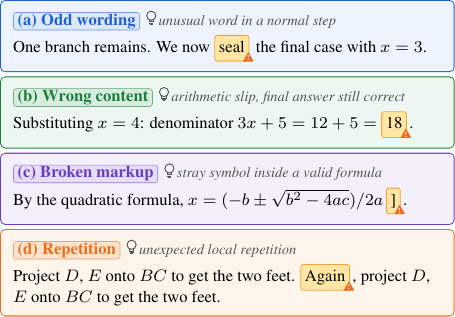}
    \caption{Examples of unreliable tokens from various sources in real correct reasoning traces.}
    \label{fig:taco_overview}
\end{figure}

Correct reasoning traces can contain locally unreliable tokens even when
the final answer is correct. This occurs because erroneous calculations,
irrelevant detours, or malformed continuations may be corrected or bypassed
before the final response. Prior work on process supervision and step-level
verification shows that outcome-based feedback can overlook flawed intermediate
reasoning~\citep{uesato2022solving,lightman2023verify}. In addition, models may
produce repeated or incoherent text inside generated traces without necessarily
disrupting the overall logical flow~\citep{holtzman2020curious,welleck2020unlikelihood}. We observe such locally unreliable tokens in real correct reasoning traces, as illustrated by the case studies in Figure~\ref{fig:taco_overview}.

The phenomenon that LLMs can produce locally unreliable tokens has been widely
studied in the decoding literature, typically through the low-probability tail
of auto-regressive policies. Prior work identifies an unreliable subset of this
region, where sampled continuations can become incoherent or off-topic; we refer to tokens from this locally implausible subset as \textbf{implausible tail tokens}~\citep{holtzman2020curious}. Standard inference-time strategies such as nucleus sampling suppress these tokens by
truncating the tail, which has been shown to improve generation
quality.

In GRPO-style training, however, full-vocabulary rollouts allow implausible tail tokens to enter the
trajectories used for policy updates. While implausible tail tokens in failed completions can be penalized by negative advantages, they become problematic when they occur in completions with positive advantages, since the broadcast rule assigns them the same positive credit as contextually sound tokens. We formalize this as a critical failure mode of the broadcast rule, which we refer to as \textbf{\textit{positive-credit contamination}}. The underlying issue is a mismatch between trajectory-level success and local token validity, as outcome rewards
assess only the final answer.

\begin{figure}[t]
    \centering
    \includegraphics[width=\columnwidth]
        {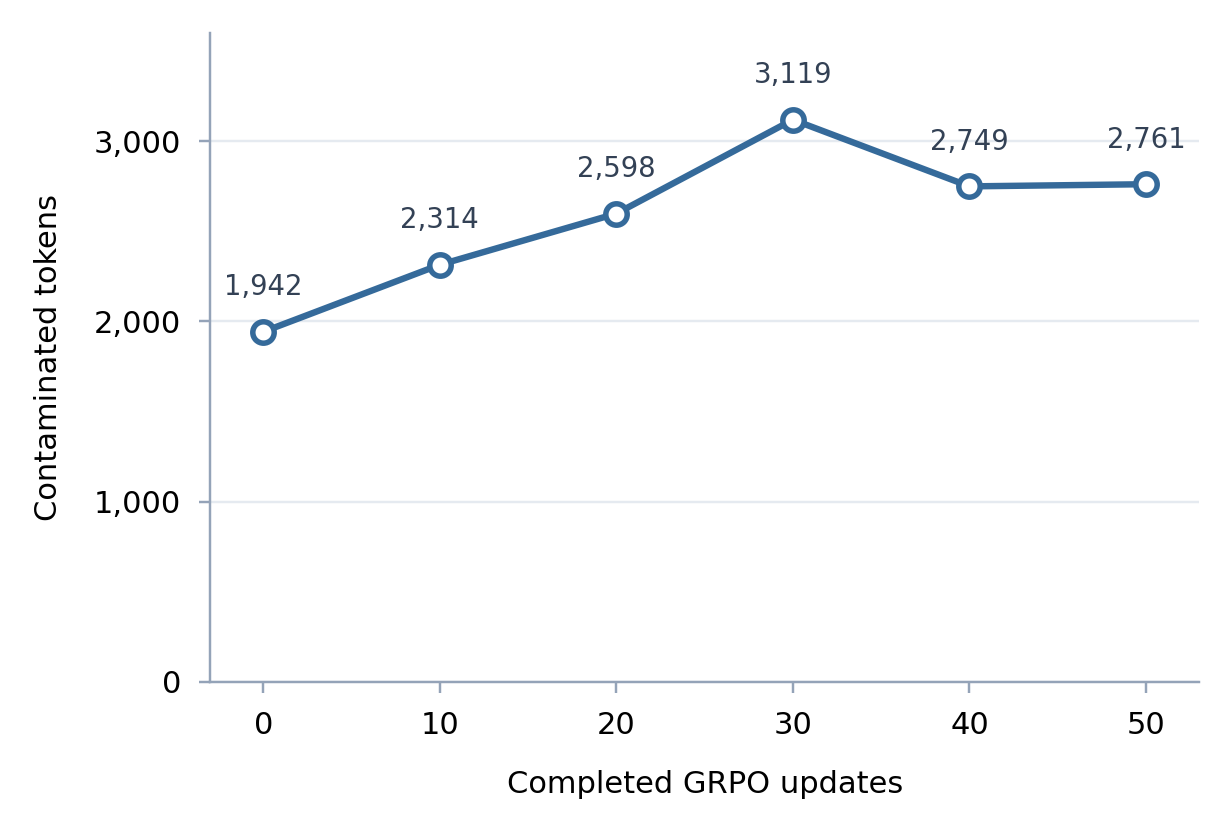}
    \caption{Number of contaminated tokens across GRPO training.}
    \label{fig:pcc_preliminary}
\end{figure}

\paragraph{Persistence during GRPO training.} To examine whether this mismatch persists during optimization,
we use Claude Opus 5.5~\cite{anthropic2026opus55}
to annotate local errors in positive-advantage rollouts
from Qwen3-1.7B-Base over the first 50 GRPO updates.
Figure~\ref{fig:pcc_preliminary} shows that contaminated
tokens, defined as tokens within annotated error spans,
persist across checkpoints, with counts ranging from
1,942 to 3,119 in the analyzed rollouts.
These tokens continue to inherit positive credit under
the broadcast rule throughout the training window.

To further demonstrate how this credit mismatch affects
learning, we conduct a controlled experiment in a synthetic
sequential Markov Decision Process (MDP), where optimal
and suboptimal actions are explicitly defined.
We compare broadcast trajectory-level advantages with
credit assignment based on each step's own reward.

\begin{figure*}[t]
    \centering
    \begin{subfigure}[t]{0.329\textwidth}
        \centering
        \includegraphics[width=1\linewidth]{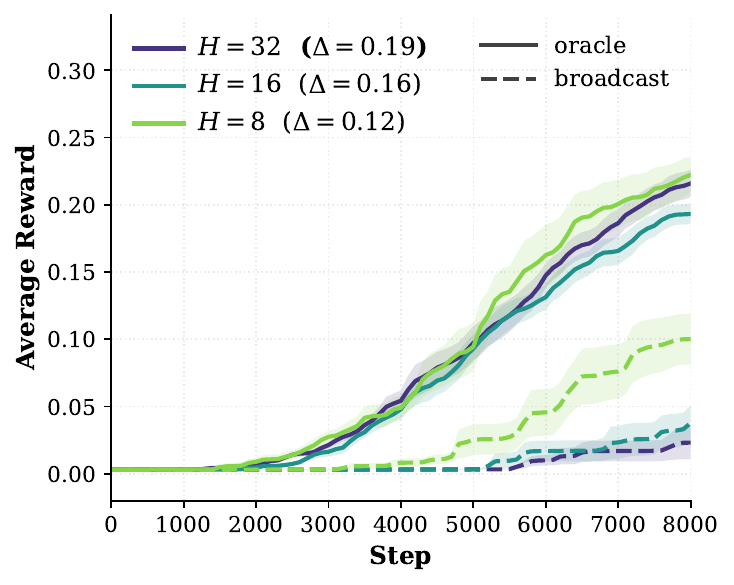}
        \caption{Trace length $H$}
    \end{subfigure}
    \hfill
    \begin{subfigure}[t]{0.329\textwidth}
        \centering
        \includegraphics[width=1\linewidth]{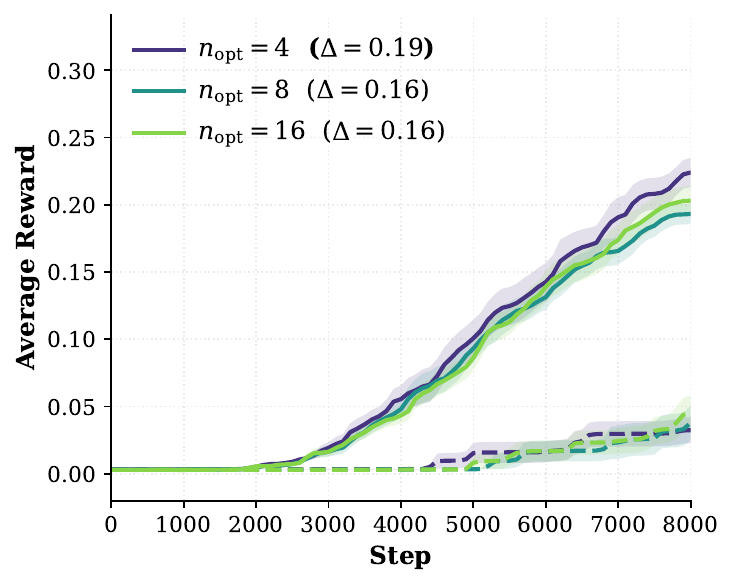}
        \caption{Optimal-action count $n_{\mathrm{opt}}$}
    \end{subfigure}
    \hfill
    \begin{subfigure}[t]{0.329\textwidth}
        \centering
        \includegraphics[width=1\linewidth]{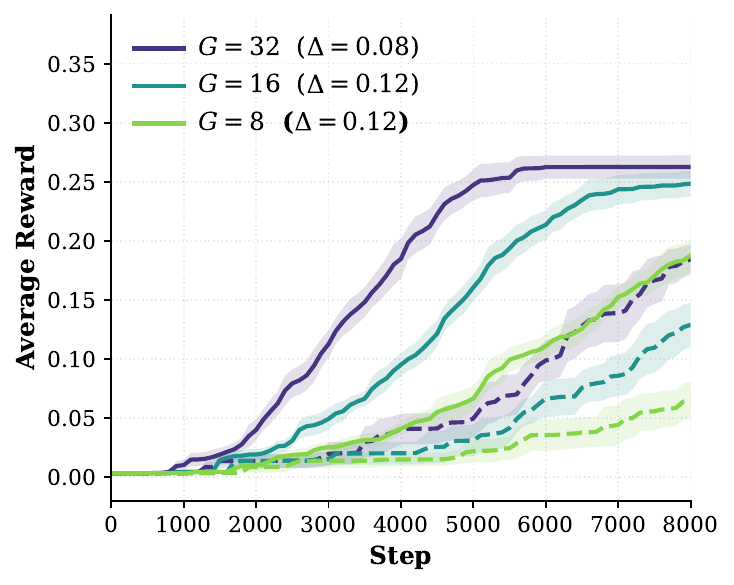}
        \caption{Group size $G$}
    \end{subfigure}
    \caption{Demonstration of the effect of positive-credit contamination on a synthetic sequential MDP. Experiments are conducted to analyze how (a) trace length $H$, (b) optimal-action count $n_{\mathrm{opt}}$, and (c) group size $G$ affect the contamination behavior. $\Delta$ denotes the final-step reward gap between oracle and broadcast.}
    \label{fig:toy_mdp}
\end{figure*}

We study three axes that mirror LLM reasoning training:
trace length $H$, optimal-action sparsity controlled by
$n_{\mathrm{opt}}$, and group size $G$.
As shown in Figure~\ref{fig:toy_mdp}, step-level credit
assignment outperforms the broadcast rule
across all settings, and the gap widens with longer traces,
sparser optimal actions, and smaller rollout groups.
These conditions reflect challenges in reasoning-oriented
RLVR, including long chain-of-thought generation,
scarce optimal continuations, and limited rollout budgets.
Together, these findings illustrate how trajectory-level
credit broadcasting can hinder learning when local actions
differ in quality, motivating token-level credit calibration.
The detailed experimental setup and results are provided
in Appendix~\ref{sec:toy_mdp}.

\section{Methodology}

In this section, we introduce \Algname{} (\Algnameabbr{})
to calibrate the positive credit each token inherits from
its trajectory and mitigate positive-credit contamination.
\Algnameabbr{} consists of two modules: policy-relative
tail-risk estimation, which computes a score measuring
a sampled token's rarity relative to the policy's local
expectations; and tail-aware credit calibration, which uses this score
to determine how much inherited positive credit to retain.
Together, these modules selectively reduce positive updates
while preserving opportunities to learn useful rare patterns.

\subsection{Policy-Relative Tail-Risk Estimation}

Mitigating positive-credit contamination requires
identifying locally problematic tokens within rewarded
completions, such that their inherited positive credit can
be selectively reduced.
However, directly labeling these errors throughout RL
training would incur prohibitive annotation costs.
\Algnameabbr{} instead utilizes the policy as its own reference:
a sampled token that is far rarer than the policy typically expects in the same context provides a signal of a potential local error and thus serves as a candidate for credit reduction.
\Algnameabbr{} quantifies how unusual each sampled token is
with a tail-risk score and uses it to determine how much
positive credit to retain.

To estimate tail risk, \Algnameabbr{} examines the local
next-token distribution at each generation step.
A low-probability token alone is not a sufficient indicator
of a local error, since it may reflect exploration when
the model is uncertain about future generations.
However, when the model is confident, the same low probability means the sampled token contradicts a strong prediction, which is far more indicative of a local error.
Therefore, \Algnameabbr{} assesses this signal relative to the policy's own expectations by comparing the sampled token's log
probability with the expected log probability in the same
context. Formally, for a sampled token $o_{i,t}$, \Algnameabbr{} defines its tail-risk score
$r^{\mathrm{tail}}_{i,t}$ through an expected log ratio:
\begin{equation}
    r^{\mathrm{tail}}_{i,t}
    =
    \mathbb{E}_{v \sim \pi_\theta(\cdot \mid c_{i,t})}
    \left[
        \log \frac{\alpha\, \pi_\theta(v \mid c_{i,t})}{p_{i,t}}
    \right].
\label{eq:tail_score_raw}
\end{equation}
A higher tail-risk score indicates that the sampled token
is more unexpected under the local policy and is treated
as a stronger signal of a potential local error.
The parameter $\alpha \in (0,1)$ controls the threshold
for identifying tail tokens. Tokens with positive scores are flagged as risky, and a larger $\alpha$ flags more of them.

\subsection{Tail-Aware Credit Calibration}
The tail-risk score provides a policy-relative signal
for moderating inherited positive credit. Based on this, \Algnameabbr{} assigns each token a risk-dependent weight that softly reduces its positive credit as the risk increases. This smooth down-weighting preserves partial gradients for genuinely useful low-probability patterns, allowing them to accumulate reinforcement across rewarded trajectories, while progressively dampening accidental noise. Formally, \Algnameabbr{} defines the token weight $w_{i,t}$ as:
\begin{equation}
    w_{i,t}
    =
    \begin{cases}
        1-\lambda\bigl(1-e^{-r^{\mathrm{tail}}_{i,t}}\bigr),
        & r^{\mathrm{tail}}_{i,t}>0, \\[3pt]
        1, & r^{\mathrm{tail}}_{i,t}\le 0.
    \end{cases}
    \label{eq:credit_weight}
\end{equation}
Tokens with non-positive risk scores retain full weight, while high-risk tokens receive smoothly decaying weights, bounded below by $1-\lambda$, where $\lambda \in (0,1)$ is
a hyperparameter controlling the maximum suppression strength. \Algnameabbr{} then applies this weight to the broadcast advantage to produce a calibrated token-level advantage:
\begin{equation}
    \hat{A}^{\Algnameabbr{}}_{i,t}
    =
    w_{i,t}^{\mathbb{I}[\hat{A}_i>0]}\hat{A}_i .
    \label{eq:calibrated_advantage}
\end{equation}
Only positive advantages are modulated; negative advantages are left
unchanged to preserve the suppression signal from failed
trajectories.
\subsection{Algorithm of \Algnameabbr{}}
\label{sec:algo}
\Algnameabbr{} integrates into the standard GRPO training loop by replacing the broadcast advantage $\hat{A}_i$ with the calibrated token-level credit $\hat{A}^{\Algnameabbr{}}_{i,t}$ from Eq.~\eqref{eq:calibrated_advantage}, while all other components of the GRPO surrogate remain unchanged. The calibration weights are computed from the current policy $\pi_\theta$ at each actor forward pass and detached before backpropagation, thus introducing negligible overhead. Algorithm~\ref{alg:taco} summarizes the full procedure.

\makeatletter
\providecommand{\PState}[1]{\State\parbox[t]{\dimexpr\linewidth-\ALG@thistlm\relax}{\raggedright\strut#1\strut}}
\makeatother
\providecommand{\algmark}[1]{\tikz[remember picture,overlay]\coordinate(#1);}
\providecommand{\algband}{\tikz[remember picture,overlay]{%
  \coordinate (tacoBR) at (tacoR |- tacoB);
  \fill[blue!60, fill opacity=0.08, rounded corners=2pt]
    ([xshift=-1.9em,yshift=1.05em]tacoL) rectangle ([xshift=1pt,yshift=-0.55em]tacoBR);}}
\providecommand{\algsec}[2]{\Statex\hspace{\algorithmicindent}{\color{#1}\textit{\#\# #2}}}

\begin{algorithm}[t]
\small
\caption{\small\Algnameabbr{}: Tail-Aware Credit Calibration}
\label{alg:taco}
\raggedright
\begin{algorithmic}[1]
\Require Prompt dataset $\mathcal{D}$; policy $\pi_\theta$; verifier $\mathcal{R}$;
group size $G$; hyperparameters $\alpha,\lambda$
\For{each training iteration}
    \algsec{gray}{Standard GRPO rollout}
    \State $\pi_{\theta_{\mathrm{old}}} \leftarrow \pi_\theta$
    \State Sample prompt batch $\mathcal{B}\sim\mathcal{D}$
    \PState{Generate $G$ completions for each prompt using $\pi_{\theta_{\mathrm{old}}}$}
    \PState{Compute rewards and group-normalized advantages $\hat{A}_i$
      \hfill{\color{gray}$\triangleright$ \textit{sequence-level credit}}}
    \vspace{3pt}
    \Statex\algmark{tacoL}\hspace{\algorithmicindent}{\color{blue!70!black}\textit{\#\# Tail-aware credit calibration}}\hfill\algmark{tacoR}
    \PState{Compute tail risk $r^{\mathrm{tail}}_{i,t}$ by Eq.~\eqref{eq:tail_score_raw}}
    \PState{Convert tail risk into credit-suppression weights $w_{i,t}$ by Eq.~\eqref{eq:credit_weight}}
    \PState{Calibrate positive sequence-level credit into token-level advantages $\hat{A}^{\Algnameabbr{}}_{i,t}$ by Eq.~\eqref{eq:calibrated_advantage}\algmark{tacoB}\algband}
    \vspace{3pt}
    \algsec{gray}{Standard GRPO policy update}
    \PState{Update $\pi_\theta$ with the GRPO clipped surrogate using $\hat{A}^{\Algnameabbr{}}_{i,t}$
      \hfill{\color{blue!70!black}$\triangleright$ \textit{token-level credit}}}
\EndFor
\end{algorithmic}
\end{algorithm}

\section{Experiments}

In this section, we conduct experiments to verify the effectiveness of our method, aiming to answer the following research questions: \textbf{RQ1:} Does \Algnameabbr{} improve performance across diverse reasoning benchmarks? \textbf{RQ2:} Does \Algnameabbr{} sustain performance gains
under longer RL training and mitigate late-stage collapse
by calibrating positive credit for persistent tail tokens? \textbf{RQ3:} Does \Algnameabbr{} selectively reduce
positive credit for questionable local choices while
preserving the learning signal for coherent reasoning
within successful trajectories?

\subsection{Experimental Setup}
\label{sec:setup}
\paragraph{Models.} We evaluate \Algnameabbr{} on four LLMs from different model
families and scales: Qwen3-1.7B-Base, Qwen3-4B-Base~\citep{yang2025qwen3}, Qwen2.5-Math-7B~\citep{yang2024qwen25math} and Olmo-3-1025-7B~\citep{olmo2025olmo3}. For the main experiments, we use DAPO-Math-17K\footnote{%
\url{https://huggingface.co/datasets/BytedTsinghua-SIA/DAPO-Math-17k}} as the training dataset. Our codebase is built on \texttt{verl}~\citep{verl2024}, and we follow its standard GRPO training recipe. All methods share the same configuration, with method-specific
hyperparameters set to the values reported in their original papers.

\paragraph{Datasets.} We evaluate the trained models on six mathematical reasoning benchmarks:
AIME 2024~\citep{aime2024}, AIME 2025~\citep{aime2025},
AMC 2023~\citep{amc2023}, MATH-500~\citep{lightman2023verify},
Minerva Math~\citep{lewkowycz2022solving}, and
OlympiadBench~\citep{he2024olympiadbench}. To assess
out-of-distribution generalization beyond mathematics, we further
evaluate on two scientific reasoning benchmarks:
MMLU-Pro~\citep{wang2024mmlupro} and
GPQA-Diamond~\citep{rein2023gpqa}. We report avg@32 for AIME and
AMC, avg@4 for MATH-500, Minerva Math, and OlympiadBench, and
avg@16 for MMLU-Pro and GPQA-Diamond.

\paragraph{Baselines.} We compare \Algnameabbr{} against five baselines. \textbf{GRPO} is the standard critic-free RLVR baseline; we implement it with clip-higher asymmetric clipping to enable long-term training~\citep{yu2025dapo}. \textbf{GRPO w/ Adv. Reweighting} reweights low-probability tokens to reduce their over-dominance in policy-gradient updates~\citep{yang2025lowprobdominate}. \textbf{GRPO w/ Forking Tokens} restricts policy-gradient updates to high-entropy tokens that steer reasoning pathways~\citep{wang2025beyond8020}.
\textbf{GTPO} assigns entropy-weighted token-level rewards to enable fine-grained credit assignment~\citep{gtpo2025}. \textbf{STAPO} dampens updates from tokens with disproportionately large gradients for more stable optimization~\citep{liu2026stapo}. Additional details are
provided in Appendix~\ref{sec:implementation_details}.
\newcommand{\best}[1]{\textbf{#1}}
\newcommand{\second}[1]{\underline{#1}}
\newcommand{\inc}[1]{{%
  \fontsize{9}{13}\selectfont
  \textcolor{green!45!black}{\textbf{(+#1)}}}}
\newcommand{\dec}[1]{{%
  \fontsize{9}{13}\selectfont
  \textcolor{red!70!black}{\textbf{(-#1)}}}}
\newcommand{\NA}{--}
\newcommand{\deltanum}[1]{{\fontsize{11}{13}\selectfont #1}}

\providecommand{\std}[1]{{\scriptsize$\pm$#1}}

\begin{table*}[t]
\centering
\small
\setlength{\tabcolsep}{2.5pt}
\renewcommand{\arraystretch}{1.12}
\resizebox{\textwidth}{!}{%
\begin{tabular}{@{}lccccccccc@{}}
\toprule
\textbf{Method}
& \textbf{AIME24}
& \textbf{AIME25}
& \textbf{AMC23}
& \textbf{MATH-500}
& \textbf{Minerva}
& \textbf{Olympiad}
& \textbf{MMLU-Pro}
& \textbf{GPQA-D}
& \textbf{Avg.} \\
\midrule

\multicolumn{10}{c}{\textbf{Qwen3-1.7B-Base}} \\
\midrule
GRPO
& 9.55\std{1.54}
& 7.53\std{0.32}
& \underline{46.85\std{1.32}}
& 65.50\std{0.43}
& 25.43\std{0.70}
& 29.01\std{1.09}
& 24.71\std{0.99}
& 20.45\std{0.44}
& 28.63\std{0.30} \\
GRPO w/ Adv. Reweighting
& 10.80\std{0.51}
& 8.30\std{0.75}
& 45.96\std{1.58}
& 64.70\std{0.65}
& \underline{25.61\std{1.38}}
& 29.19\std{1.22}
& \underline{29.72\std{0.44}}
& \underline{22.64\std{1.09}}
& \underline{29.62\std{0.61}} \\
GRPO w/ Forking Tokens
& 11.53\std{0.51}
& 7.08\std{0.63}
& 44.30\std{1.36}
& 67.20\std{0.13}
& \textbf{\boldmath 25.74\std{1.19}}
& 28.18\std{0.56}
& 23.83\std{1.20}
& 21.83\std{0.49}
& 28.71\std{0.27} \\
GTPO
& 10.73\std{0.45}
& 8.51\std{1.07}
& 46.61\std{1.32}
& 65.07\std{0.63}
& 24.94\std{2.13}
& 29.08\std{0.19}
& 23.67\std{0.52}
& 22.00\std{0.60}
& 28.83\std{0.79} \\
STAPO
& \underline{11.94\std{0.69}}
& \textbf{\boldmath 9.72\std{1.41}}
& 46.64\std{0.55}
& \underline{69.65\std{0.94}}
& 23.53\std{2.20}
& \underline{30.28\std{1.94}}
& 24.33\std{0.53}
& 18.31\std{1.03}
& 29.30\std{0.36} \\
\textbf{\Algnameabbr{} (Ours)}
& \textbf{\boldmath 14.90\std{1.20}}
& \underline{8.92\std{1.30}}
& \textbf{\boldmath 49.09\std{1.33}}
& \textbf{\boldmath 70.05\std{0.13}}
& 25.00\std{0.97}
& \textbf{\boldmath 31.39\std{1.06}}
& \textbf{\boldmath 30.78\std{1.06}}
& \textbf{\boldmath 24.54\std{0.51}}
& \textbf{\boldmath 31.83\std{0.47}} \\
\hspace{1.2em}$\Delta$ vs. GRPO
& \inc{5.35}
& \inc{1.39}
& \inc{2.24}
& \inc{4.55}
& \dec{0.43}
& \inc{2.38}
& \inc{6.07}
& \inc{4.09}
& \inc{3.20} \\

\midrule
\multicolumn{10}{c}{\textbf{Qwen3-4B-Base}} \\
\midrule
GRPO
& \underline{26.49\std{1.15}}
& 19.79\std{1.45}
& 67.81\std{0.21}
& 76.98\std{0.93}
& 34.93\std{0.32}
& 36.07\std{1.45}
& 36.10\std{1.08}
& 26.23\std{0.67}
& 40.55\std{0.33} \\
GRPO w/ Adv. Reweighting
& 24.48\std{0.68}
& 21.04\std{0.73}
& \underline{70.96\std{0.37}}
& \underline{77.73\std{0.63}}
& \underline{35.11\std{1.22}}
& \underline{37.77\std{1.11}}
& \underline{39.12\std{0.88}}
& \underline{26.96\std{0.50}}
& \underline{41.65\std{0.25}} \\
STAPO
& 25.00\std{1.20}
& \underline{22.60\std{1.35}}
& \textbf{\boldmath 72.03\std{1.02}}
& 76.57\std{0.23}
& 34.22\std{1.25}
& 37.15\std{0.58}
& 38.85\std{0.66}
& 26.45\std{1.44}
& 41.61\std{0.08} \\
\textbf{\Algnameabbr{} (Ours)}
& \textbf{\boldmath 27.01\std{0.30}}
& \textbf{\boldmath 24.90\std{2.57}}
& 70.34\std{1.77}
& \textbf{\boldmath 80.77\std{0.76}}
& \textbf{\boldmath 35.81\std{0.54}}
& \textbf{\boldmath 39.91\std{1.95}}
& \textbf{\boldmath 42.51\std{0.42}}
& \textbf{\boldmath 29.08\std{1.10}}
& \textbf{\boldmath 43.79\std{0.29}} \\
\hspace{1.2em}$\Delta$ vs. GRPO
& \inc{0.52}
& \inc{5.11}
& \inc{2.53}
& \inc{3.79}
& \inc{0.88}
& \inc{3.84}
& \inc{6.41}
& \inc{2.85}
& \inc{3.24} \\

\midrule
\multicolumn{10}{c}{\textbf{Qwen2.5-Math-7B}} \\
\midrule
GRPO
& \underline{29.93\std{0.66}}
& 17.19\std{0.72}
& 71.56\std{0.79}
& 82.97\std{1.37}
& 50.89\std{2.21}
& 47.16\std{0.59}
& \underline{29.78\std{0.82}}
& 23.28\std{0.30}
& 44.09\std{0.29} \\
GRPO w/ Adv. Reweighting
& 28.58\std{1.36}
& 16.35\std{0.58}
& \underline{75.86\std{1.11}}
& 82.30\std{0.69}
& 50.34\std{1.47}
& 47.27\std{1.60}
& 29.69\std{0.19}
& 24.04\std{0.30}
& 44.30\std{0.54} \\
STAPO
& 28.85\std{1.82}
& \underline{17.71\std{0.48}}
& 72.45\std{1.45}
& \textbf{\boldmath 85.92\std{0.47}}
& \underline{53.34\std{0.61}}
& \underline{47.95\std{1.74}}
& 27.35\std{0.85}
& \textbf{\boldmath 25.61\std{0.48}}
& \underline{44.90\std{0.28}} \\
\textbf{\Algnameabbr{} (Ours)}
& \textbf{\boldmath 30.56\std{1.26}}
& \textbf{\boldmath 19.13\std{0.57}}
& \textbf{\boldmath 78.85\std{1.25}}
& \underline{83.75\std{1.85}}
& \textbf{\boldmath 54.01\std{1.25}}
& \textbf{\boldmath 49.38\std{1.35}}
& \textbf{\boldmath 30.22\std{0.68}}
& \underline{24.49\std{1.34}}
& \textbf{\boldmath 46.30\std{0.33}} \\
\hspace{1.2em}$\Delta$ vs. GRPO
& \inc{0.63}
& \inc{1.94}
& \inc{7.29}
& \inc{0.78}
& \inc{3.12}
& \inc{2.22}
& \inc{0.44}
& \inc{1.21}
& \inc{2.21} \\

\midrule
\multicolumn{10}{c}{\textbf{Olmo-3-1025-7B}} \\
\midrule
GRPO
& \underline{27.22\std{1.44}}
& \underline{26.42\std{1.53}}
& \underline{69.04\std{0.67}}
& \underline{82.97\std{1.37}}
& \underline{42.98\std{1.25}}
& \underline{41.54\std{1.28}}
& \underline{33.45\std{0.24}}
& \underline{28.03\std{0.54}}
& \underline{43.96\std{0.25}} \\
\textbf{\Algnameabbr{} (Ours)}
& \textbf{\boldmath 29.72\std{1.14}}
& \textbf{\boldmath 29.48\std{1.53}}
& \textbf{\boldmath 73.91\std{1.13}}
& \textbf{\boldmath 83.97\std{1.23}}
& \textbf{\boldmath 45.77\std{0.96}}
& \textbf{\boldmath 43.83\std{1.93}}
& \textbf{\boldmath 35.38\std{0.91}}
& \textbf{\boldmath 30.68\std{0.82}}
& \textbf{\boldmath 46.59\std{0.59}} \\
\hspace{1.2em}$\Delta$ vs. GRPO
& \inc{2.50}
& \inc{3.06}
& \inc{4.87}
& \inc{1.00}
& \inc{2.79}
& \inc{2.29}
& \inc{1.93}
& \inc{2.65}
& \inc{2.63} \\

\bottomrule
\end{tabular}%
}
\caption{Main results on in-domain mathematical reasoning benchmarks and
out-of-distribution scientific reasoning benchmarks.
Results are reported as mean $\pm$ standard deviation over three runs.
The best and second-best results are bolded
and underlined, respectively.
The $\Delta$ row reports score differences relative to GRPO.}
\label{tab:main_results}
\end{table*}

\subsection{Performance on Reasoning Tasks (RQ1)}
Table~\ref{tab:main_results} shows the accuracy (\%) of \Algnameabbr{} and baseline methods across all benchmarks.

\paragraph{Performance Gain.} As shown in Table~\ref{tab:main_results}, \Algnameabbr{} surpasses all baselines on average and consistently improves performance across different models and benchmarks.
These gains suggest that \Algnameabbr{} improves performance by calibrating implausible positive credit among low-probability tokens, going beyond merely stabilizing their updates as in GRPO w/ Adv.\ Reweighting or using token entropy to guide credit assignment as in GRPO w/ Forking Tokens and GTPO. Compared with STAPO's hard masking using separate probability and entropy thresholds, \Algnameabbr{} softly reduces positive credit based on policy-relative tail risk, retaining nonzero weights to support rare-token learning and helping provide a more effective optimization signal.

\paragraph{OOD Generalization.} \Algnameabbr{} demonstrates strong generalization beyond the training domain. \Algnameabbr{} maintains stable gains on MMLU-Pro and GPQA-Diamond, which
cover scientific fields including physics, chemistry, and biology. By selectively suppressing tail tokens while preserving credit for useful exploratory ones, \Algnameabbr{} avoids overly aggressive gradient suppression that could limit cross-domain
generalization. This design helps the policy develop
general reasoning capabilities rather than overfitting to
domain-specific patterns.

\paragraph{Model Scale Consistency.} \Algnameabbr{} achieves consistent improvements across model families of different parameter scales. It evaluates sampled-token surprisal against the policy's expectation within the same context, allowing positive-credit calibration to naturally adapt to each model's uncertainty profile and enabling \Algnameabbr{} to improve reasoning capability without requiring model-specific tuning.
\subsection{Analysis of Training Dynamics}

\begin{figure*}[t]
    \centering
    \begin{subfigure}[t]{0.32\textwidth}
        \centering
        \includegraphics[width=1\linewidth]{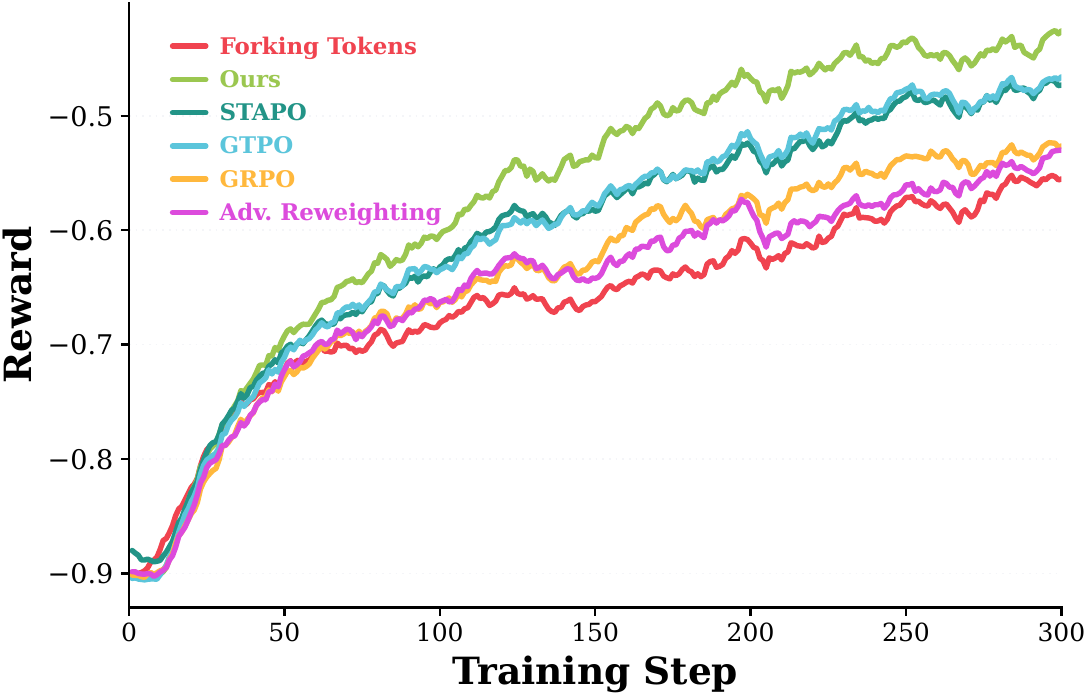}
        \caption{Training reward}
        \label{fig:a}
    \end{subfigure}
    \hfill
    \begin{subfigure}[t]{0.32\textwidth}
        \centering
        \includegraphics[width=1\linewidth]{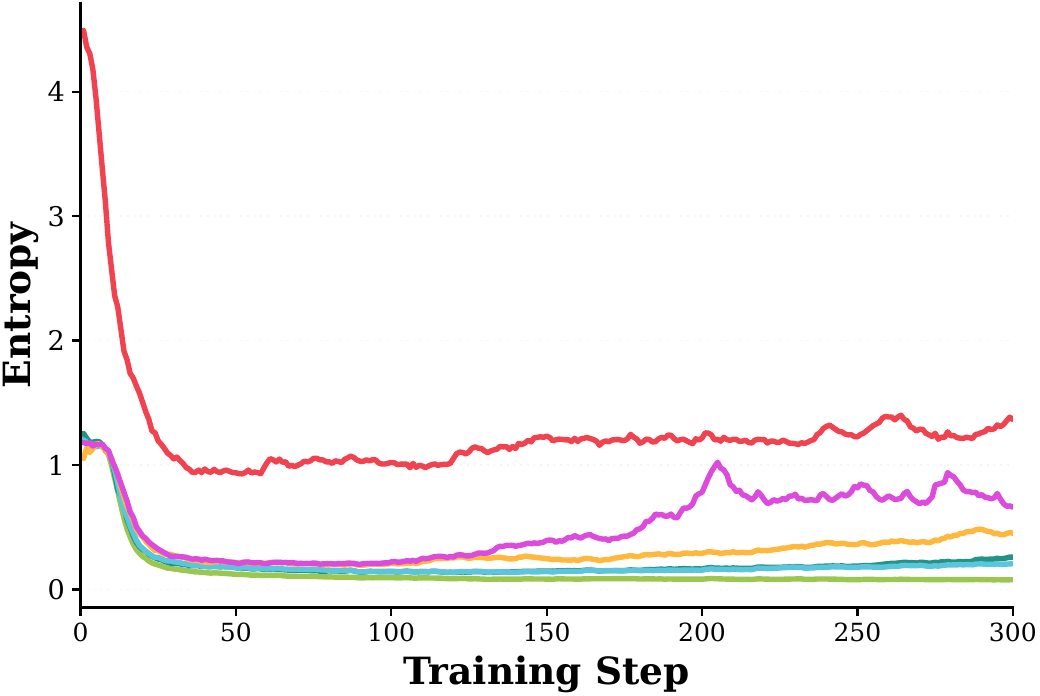}
        \caption{Policy entropy}
        \label{fig:b}
    \end{subfigure}
    \hfill
    \begin{subfigure}[t]{0.32\textwidth}
        \centering
        \includegraphics[width=1\linewidth]{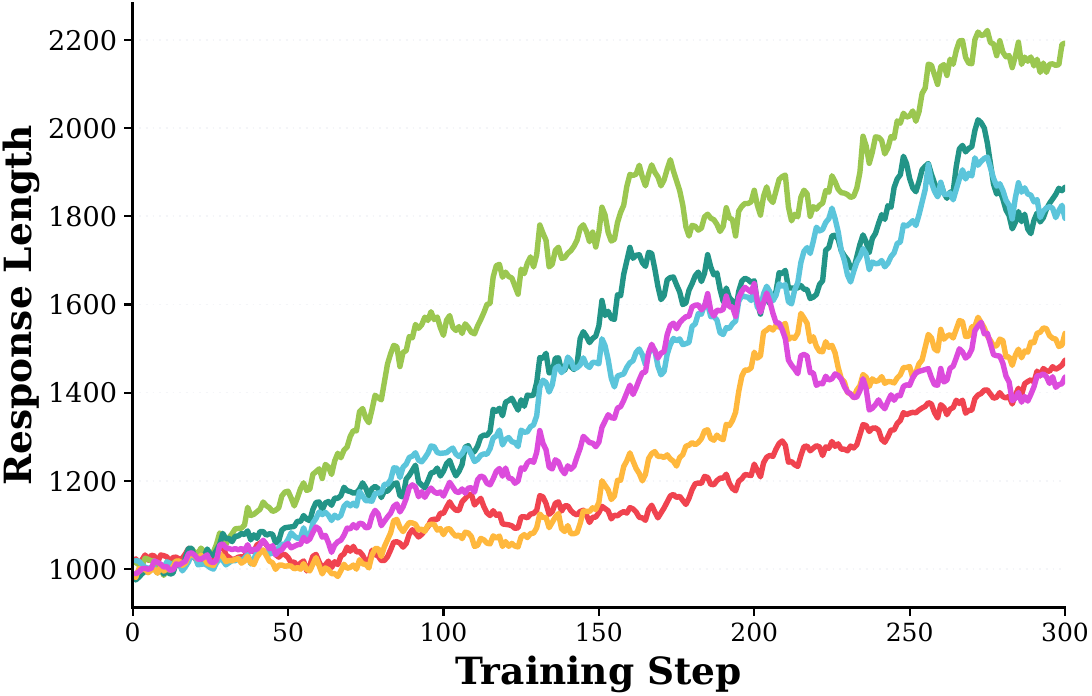}
        \caption{Response length}
        \label{fig:c}
    \end{subfigure}
    \caption{
    Training dynamics of \Algnameabbr{} and baseline methods on Qwen3-1.7B-Base.
    Results on other models are provided in Appendix~\ref{app:training_dynamics}.
    }
    \label{fig:dynamic}
\end{figure*}
\begin{figure*}[t]
    \centering
    \begin{subfigure}[t]{0.48\textwidth}
        \centering
        \includegraphics[width=\linewidth]{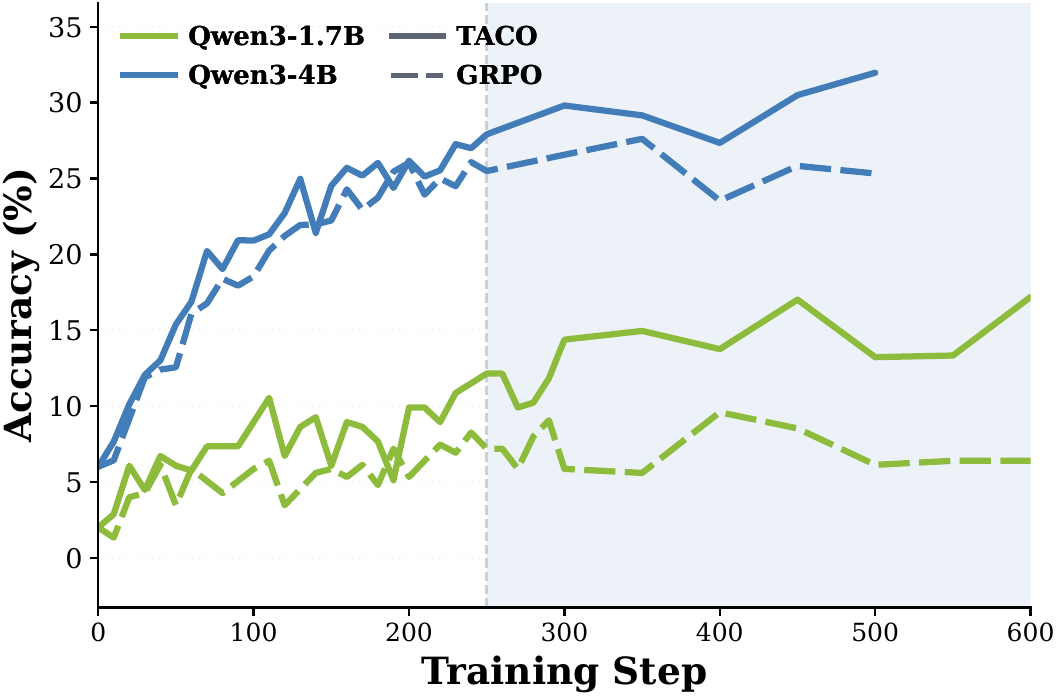}
        \caption{AIME24 test accuracy}
        \label{fig:qwen25_math_15b}
    \end{subfigure}
    \hfill
    \begin{subfigure}[t]{0.48\textwidth}
        \centering
        \includegraphics[width=\linewidth]{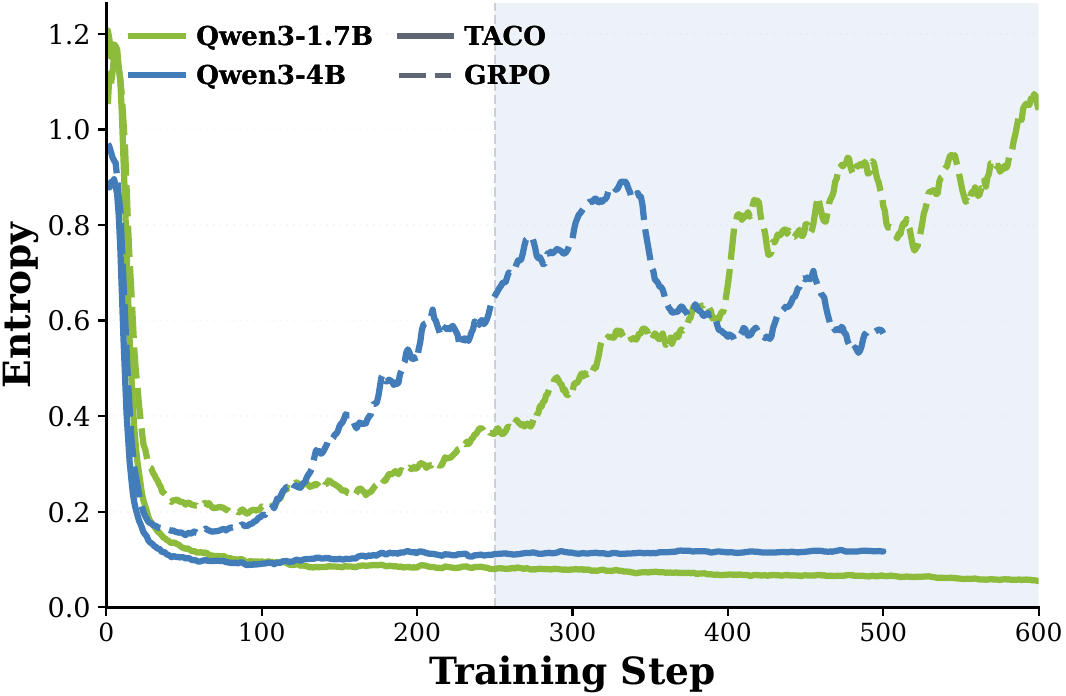}
        \caption{Policy entropy}
    \end{subfigure}
    \caption{Comparison of \Algnameabbr{} and GRPO on Qwen3-1.7B-Base and Qwen3-4B-Base with extended training. }
    \label{fig:long_horizon}
\end{figure*}

To examine how \Algnameabbr{} affects training, we compare its reward, policy entropy, and response length dynamics with baseline methods in Figure~\ref{fig:dynamic}.

\paragraph{Learning Effectiveness.} As shown in Figure~\ref{fig:a}, \Algnameabbr{} achieves
consistently higher training accuracy than baselines. This improvement stems from the selective nature of
\Algnameabbr{}'s credit calibration: by suppressing positive credit
for locally implausible tokens, each policy update concentrates
reinforcement on contextually more reliable reasoning steps,
leading to more effective policy optimization.

\paragraph{Stable Entropy and Sustained Exploration.} As shown in Figures~\ref{fig:b} and~\ref{fig:c}, \Algnameabbr{}
maintains relatively lower and more stable token entropy compared to
baselines while simultaneously producing longer responses throughout
training. The stable entropy indicates that \Algnameabbr{} suppresses
erratic tail-token updates without collapsing the policy's
exploration capacity. 

Meanwhile, the increased response length suggests that the policy
develops more complex and complete reasoning ability under this
stable optimization. Together, these dynamics suggest that \Algnameabbr{}'s credit
calibration enables the policy to maintain effective exploration
and develop more robust reasoning capabilities.
\begin{table*}[t]
\centering
\small
\setlength{\tabcolsep}{3.2pt}
\renewcommand{\arraystretch}{1.15}
\resizebox{\textwidth}{!}{%
\begin{tabular}{@{}lccccccccc@{}}
\toprule
\textbf{Method}
& \textbf{AIME24}
& \textbf{AIME25}
& \textbf{AMC23}
& \textbf{MATH-500}
& \textbf{Minerva}
& \textbf{Olympiad}
& \textbf{MMLU-Pro}
& \textbf{GPQA-D}
& \textbf{Avg.} \\
\midrule
\textbf{TACO}
& \textbf{14.38} & \textbf{9.06} & \textbf{49.45}
& \textbf{68.35} & \textbf{25.74} & \textbf{31.71}
& \textbf{30.45} & \textbf{24.43} & \textbf{31.70} \\
\midrule
Probability-only
& 11.25 & 8.44 & 47.03 & 64.75 & 24.63 & 29.11 & 29.84 & 23.36 & 29.80 \\
Entropy-only
& 10.42 & 7.60 & 46.02 & 65.15 & 24.63 & 29.08 & 23.94 & 21.37 & 28.53 \\
Random
& 8.12 & 5.73 & 39.77 & 62.30 & 24.08 & 25.63 & 26.16 & 24.43 & 27.03 \\
Hard clipping
& 10.21 & 7.19 & 47.27 & 64.75 & 25.46 & 29.56 & 26.38 & 22.19 & 29.13 \\
\bottomrule
\end{tabular}%
}
\caption{Ablation study of TACO on Qwen3-1.7B-Base.
We compare the full method with probability-only, entropy-only,
random reweighting, and hard clipping variants.}
\label{tab:ablation_mechanism}
\end{table*}

\begin{table}[t]
\centering
\small
\setlength{\tabcolsep}{3pt}
\renewcommand{\arraystretch}{1.08}
\begin{tabular}{@{}lrrrr@{}}
\toprule
& \multicolumn{2}{c}{$\alpha = 0.01$}
& \multicolumn{2}{c}{$\alpha = 0.005$} \\
\cmidrule(lr){2-3}
\cmidrule(lr){4-5}
Dataset
& $\lambda = 0.6$ & $\lambda = 0.9$
& $\lambda = 0.6$ & $\lambda = 0.9$ \\
\midrule
AIME24
& \best{15.10} & \second{14.38}
& 12.81 & 13.23 \\
AIME25
& \best{9.58} & \second{9.06}
& 8.64 & \second{9.06} \\
AMC23
& 47.50 & \best{49.45}
& \second{48.15} & 46.57 \\
MATH-500
& \second{68.15} & \best{68.35}
& 67.20 & 67.55 \\
Minerva
& 24.82 & \best{25.74}
& 24.72 & \second{25.55} \\
Olympiad
& 28.97 & \best{31.71}
& 29.38 & \second{30.08} \\
MMLU-Pro
& \best{30.95} & \second{30.45}
& 28.85 & 29.85 \\
GPQA-D
& 23.99 & 24.43
& \best{26.26} & \second{25.81} \\
\midrule
Average
& \second{31.13} & \best{31.70}
& 30.75 & 30.96 \\
\bottomrule
\end{tabular}
\caption{Sensitivity analysis on Qwen3-1.7B-Base.}
\label{tab:ablation_alpha_lambda}
\vspace{-2mm}
\end{table}
\subsection{Long-Horizon Training Stability (RQ2)}

A challenge in GRPO-style training is that the policy tends
to plateau or degrade as training progresses, limiting the benefit
of extended optimization. The persistence of local errors in positive-advantage
trajectories observed in our preliminary analysis motivates examining whether calibrating their
positive credit can sustain improvements over longer training. 

To evaluate \Algnameabbr{}'s robustness in
this regime, we double the default training budget for both GRPO and
\Algnameabbr{} on two model scales: from 300 to 600 steps on
Qwen3-1.7B-Base and from 250 to 500 steps on Qwen3-4B-Base. As shown in Figure~\ref{fig:long_horizon}, GRPO's test accuracy on AIME24 plateaus during the extended training phase and subsequently declines, while
\Algnameabbr{} continues to improve throughout the extended phase. The entropy dynamics further show that GRPO exhibits large fluctuations as training proceeds, whereas \Algnameabbr{} maintains a smooth and stable entropy profile, avoiding both entropy collapse and explosion. 

These results support the effectiveness of calibrating
positive credit for persistent tail tokens in sustaining
long-horizon performance gains and mitigating late-stage
degradation, allowing the policy to continue benefiting
from extended optimization where standard GRPO stagnates.

\subsection{Behavioral Analysis (RQ3)}
\label{sec:behavioral_analysis}
\begin{figure}[t]
    \centering
    \includegraphics[width=1\linewidth]{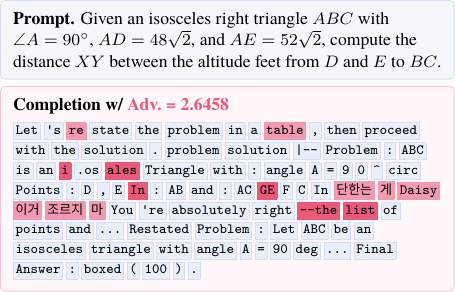}
    \caption{A case from \textbf{Qwen3-4B-Base}.
\Algnameabbr{} selectively downweights tail tokens (shown in red; darker red indicates stronger suppression), while preserving coherent reasoning steps (shown in blue).}
    \label{fig:behavioral_case}
\end{figure}

To further illustrate how \Algnameabbr{} mitigates positive-credit contamination, we present a rewarded trace sampled during training in Figure~\ref{fig:behavioral_case}. Although the completion reaches the correct answer, its early generation contains unreliable tail tokens, including unnecessary table formatting, corrupted math terms (\texttt{i}), broken formula (\texttt{In}), mixed-language noise (\texttt{Daisy}), and irrelevant text (\texttt{--the}, \texttt{list}). Under the GRPO broadcast rule, these unreliable behaviors would be mis-amplified by receiving the same positive advantage as useful reasoning tokens. In contrast, \Algnameabbr{} assigns suppressed credit to these tokens while preserving full credit for the later valid solution steps, showing that \Algnameabbr{} can help suppress harmful local behaviors without weakening coherent reasoning. Additional qualitative cases and training diagnostics are provided in Appendix~\ref{app:training_dynamics}.

\subsection{Ablation Study}
We ablate TACO's policy-relative credit calibration.
\textbf{Probability-only} removes contextual calibration and selects tokens solely by their probabilities;
\textbf{Entropy-only} uses policy entropy without considering the realized token;
\textbf{Random reweighting} randomly selects from the same positive-advantage token pool, preserving the downweighting rate and strength; \textbf{Hard clipping} uses the same tail-risk criterion
but sets the positive credit of selected tokens to zero,
testing the benefit of retaining nonzero credit. All other training settings are kept identical across variants.
Table~\ref{tab:ablation_mechanism} shows that TACO achieves
the highest average score, outperforming all
ablated variants by 1.90 to 4.67 points. These results suggest that TACO's gains depend on targeted,
policy-relative credit allocation: neither single-signal
selection nor random downweighting at the same rate reproduces its performance. The advantage over hard clipping further supports retaining partial credit, preserving learning opportunities for useful
rare patterns that recur across trajectories.

\subsection{Hyperparameter Sensitivity Study}
We evaluate the sensitivity of \Algnameabbr{} to its two key hyperparameters on Qwen3-1.7B-Base model: the tail-risk strictness $\alpha$ and the suppression strength $\lambda$. We vary $\alpha$ and $\lambda$ around the default configuration while keeping all other training settings fixed. As shown in Table~\ref{tab:ablation_alpha_lambda}, our default setting, $(\alpha,\lambda)=(0.01,0.9)$, achieves the best average performance, while \Algnameabbr{} remains effective across a reasonable range of settings. This suggests that the gains do not rely on a brittle hyperparameter choice. We also observe that setting $\alpha$ too large, e.g., $\alpha=0.1$, leads to early-stage performance degradation. This suggests that overly aggressive tail-risk identification may suppress useful low-probability behaviors together with unreliable ones.

\section{Conclusion}
In this work, we identify positive-credit contamination
as a failure mode of GRPO-style RLVR, where locally erroneous
tail tokens can inherit positive credit from successful
trajectories.
To address this issue, we propose \Algname{} (\Algnameabbr{}), which calibrates token-level credit to avoid reinforcing flawed local continuations.
\Algnameabbr{} estimates tail risk from each sampled token's rarity
relative to the policy's local expectations. It then applies a partial reduction to positive credit for high-risk tokens, limiting their reinforcement
while preserving opportunities to learn useful rare patterns.
Experiments across LLMs and reasoning benchmarks
show that \Algnameabbr{} consistently improves average
performance over GRPO-style baselines and supports
more stable long-horizon training. These results demonstrate the effectiveness of \Algnameabbr{} for improving reasoning-oriented RLVR.
\FloatBarrier
\section*{Limitations}
\label{sec:limitations}
We propose \Algnameabbr{} to mitigate positive-credit contamination in GRPO-style RLVR, and several aspects of its scope remain to be explored. First, our experiments focus on mathematical and scientific reasoning; extending \Algnameabbr{} to other verifiable domains such as code generation and tool use, or to open-ended tasks with richer reward signals, is left for future work. Second, due to computational constraints, we evaluate dense models from 1.7B to 7B parameters and do not include larger dense or mixture-of-experts models. Third, we instantiate \Algnameabbr{} within GRPO and leave other on-policy rollout paradigms, such as on-policy distillation, unexplored. As it requires only sampled tokens and the policy's next-token distribution, \Algnameabbr{} can in principle be integrated into these paradigms with minimal modification.

\bibliography{references}

\clearpage
\appendix

\section{Implementation Details}
\label{sec:implementation_details} 
\paragraph{Training Setup.}
We train all methods with vLLM-based~\cite{kwon2023efficient} rollout generation and no KL penalty. Unless otherwise specified, we use group size \(8\), learning rate \(1\times 10^{-6}\) with 10 warmup steps, maximum prompt length 1024, maximum response length 4096, rollout temperature 1.0, top-\(p=1.0\), and asymmetric clipping with lower and upper clip ranges \((0.2, 0.28)\). For the Qwen3 series, the rollout batch size is 256 and the PPO mini-batch size is 64; Qwen3-1.7B-Base and Qwen3-4B-Base are trained for 300 and 250 steps, respectively. For Qwen2.5-Math-7B, the rollout batch size is 512, the PPO mini-batch size is 32, and the model is trained for 200 steps. For Olmo3-1025-7B, the rollout batch size is 128, the PPO mini-batch size is 32, and is trained for 150 steps.

\paragraph{Baselines and Evaluation.}
For method-specific hyperparameters, TACO uses $\alpha=0.01$ and $\lambda=0.9$.
For GRPO with Advantage Reweighting, we set the reweighting coefficient to $0.1$.
For GTPO, we set the base reward coefficient to $1.0$ and the entropy bonus coefficient to $0.1$.
For Forking Tokens, we retain the top $20\%$ highest-entropy tokens within each batch for policy-gradient updates.
For STAPO, we set the selected token ratio to $20\%$ and the threshold to $0.002$.
During evaluation, we use temperature $0.7$, top-$p=0.9$, and a maximum generation length of 16{,}384 tokens for all models and benchmarks.
We follow existing work~\cite{xu2025dts, chen2025verithinker, xu2025ensemble, xu2026learning} to construct evaluation prompts, extract final answers, and build validation sets.
Specifically, we instruct models to ``Please reason step by step, and put your final answer within \texttt{\textbackslash boxed\{\}}''.
For mathematical benchmarks, we compare the final boxed answer with the reference using normalized exact match or numerical equality.
For GPQA-D and MMLU-Pro, a response is counted as correct only if its final boxed answer specifies the correct option letter, e.g., \texttt{\textbackslash boxed\{B\}}.
This format-sensitive protocol is applied identically to all methods and evaluates both reasoning correctness and adherence to the required answer format.
For final reporting, we select the checkpoint with the highest validation accuracy for each method and model.
\section{Controlled Experiment}
\label{sec:toy_mdp}

In the sequential MDP, each trajectory consists of multiple independent steps with local 
step rewards, while the trajectory-level return is the sum of these rewards. 
This design allows us to isolate credit contamination by comparing two credit 
assignment rules under the same rollout distribution and reward function: the 
broadcast rule, which assigns the group-normalized trajectory advantage to 
every step, and step-level credit assignment, which assigns credit using 
each step's own reward. This comparison illustrates how trajectory-level credit
broadcasting can affect learning when local actions
differ in quality.

\paragraph{Task Setting.}
We instantiate the synthetic sequential MDP described in 
Section~\ref{sec:pcc} as follows. Each trajectory has length \(H\). At each step, the agent samples from a large action space \(|\mathcal A|=10^5\). Among them, \(n_{\mathrm{opt}}\) actions are optimal with reward \(1\) and all remaining actions receive reward \(0\). The trajectory return is the sum of step rewards, and we report the normalized return by dividing it by \(H\). The policy is a per-step tabular softmax \(\theta\in\mathbb R^{H\times |\mathcal A|}\). To mimic a pretrained non-uniform prior, logits of optimal actions are initialized from \(\mathcal N(1,1)\), while all other logits are initialized from \(\mathcal N(0,1)\).

\paragraph{Hyper-parameters.}
The policy is trained with SGD using learning rate \(0.2\) for \(8000\) steps, with evaluation every \(100\) steps. We vary \(H\), \(n_{\mathrm{opt}}\), and group size \(G\) to study the effects of trace length, optimal-action sparsity, and group-level credit noise. All curves are averaged over ten random seeds.

\section{Additional Training Dynamics and Analysis}

\begin{figure*}[t]
    \centering
    \begin{subfigure}[t]{0.32\textwidth}
        \centering
        \includegraphics[width=1\linewidth]{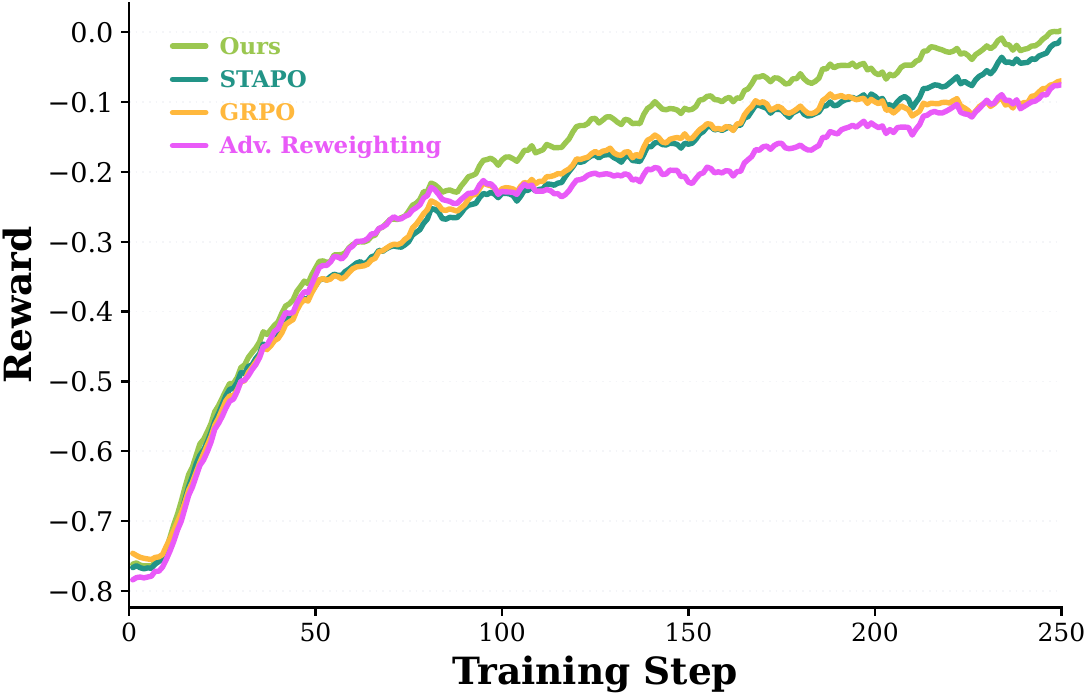}
        \caption{Training reward}
        \label{fig:4a}
    \end{subfigure}
    \hfill
    \begin{subfigure}[t]{0.32\textwidth}
        \centering
        \includegraphics[width=1\linewidth]{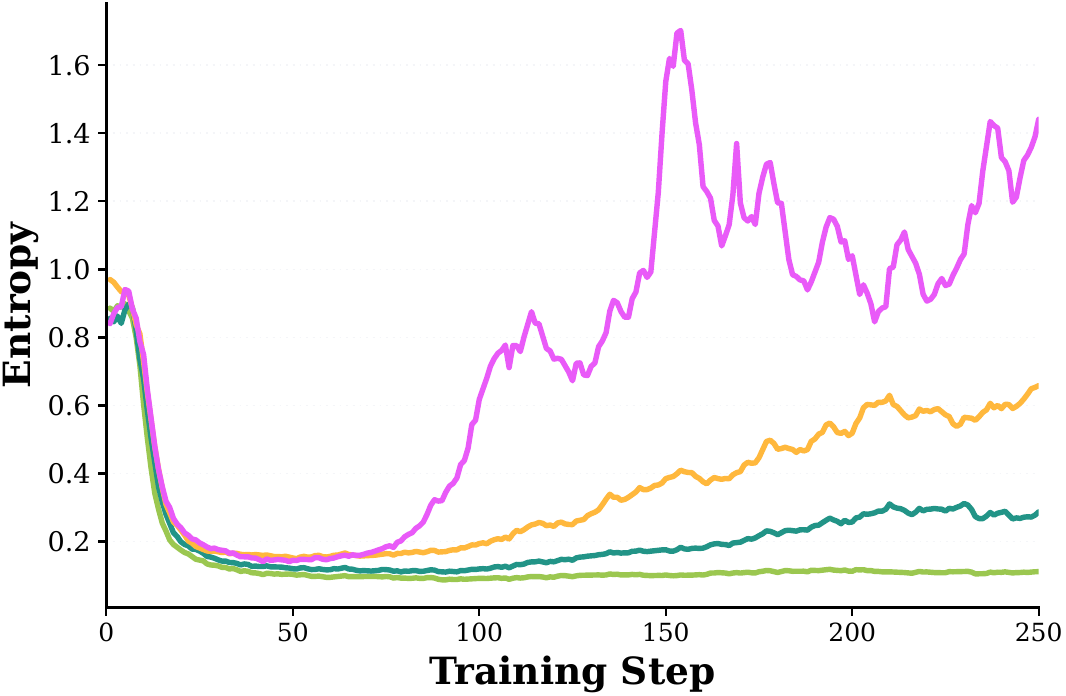}
        \caption{Policy entropy}
        \label{fig:4b}
    \end{subfigure}
    \hfill
    \begin{subfigure}[t]{0.32\textwidth}
        \centering
        \includegraphics[width=1\linewidth]{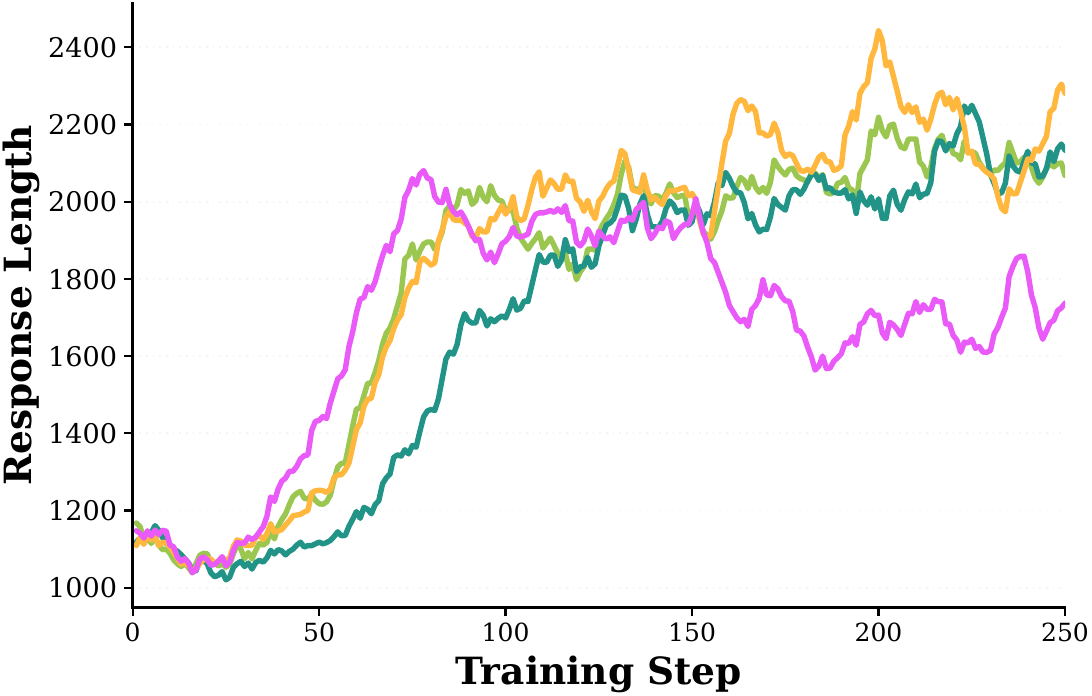}
        \caption{Response length}
        \label{fig:4c}
    \end{subfigure}
    \caption{
    Training dynamics of \Algnameabbr{} and baseline methods on Qwen3-4B-Base. 
    }
    \label{fig:dynamic1}
\end{figure*}

\begin{figure*}[t]
    \centering
    \begin{subfigure}[t]{0.32\textwidth}
        \centering
        \includegraphics[width=1\linewidth]{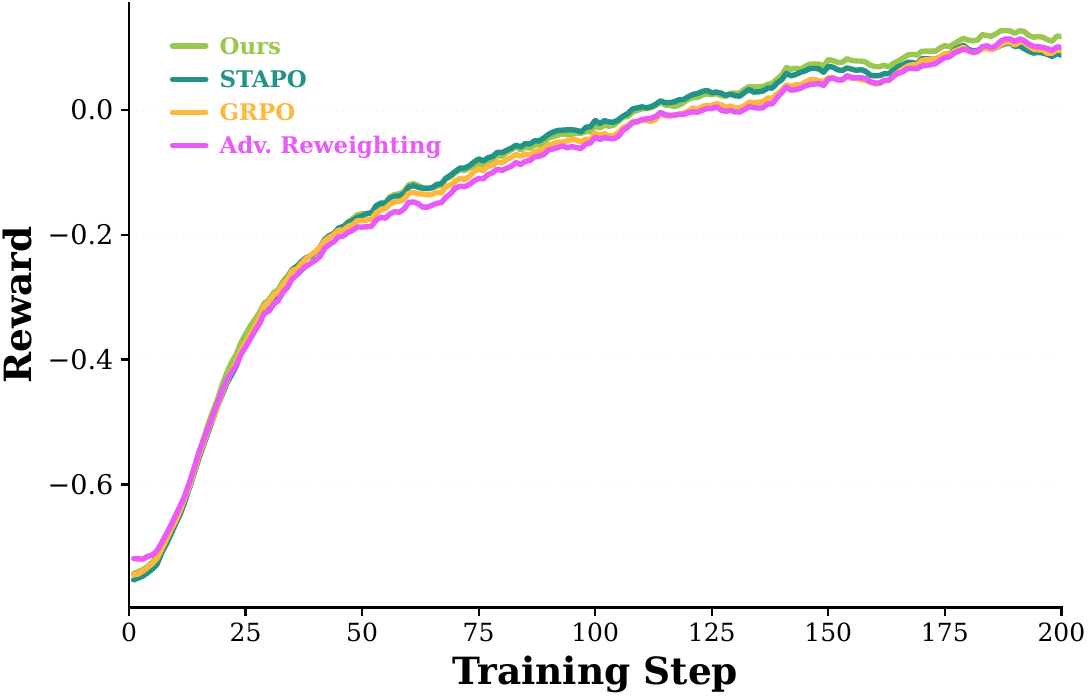}
        \caption{Training reward}
        \label{fig:7a}
    \end{subfigure}
    \hfill
    \begin{subfigure}[t]{0.32\textwidth}
        \centering
        \includegraphics[width=1\linewidth]{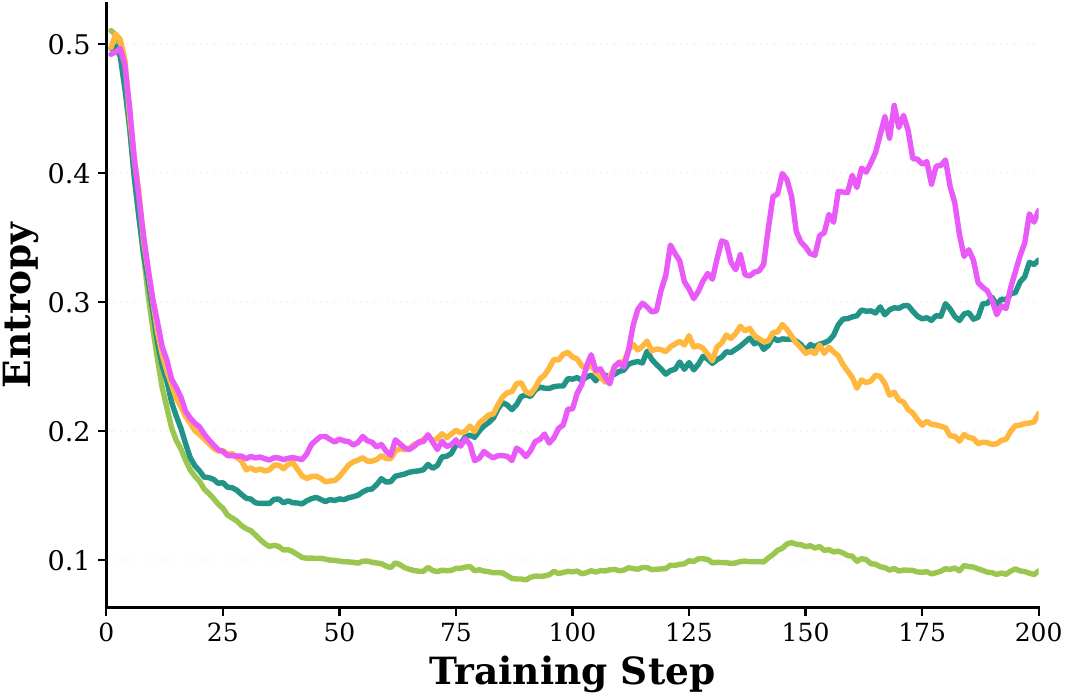}
        \caption{Policy entropy}
        \label{fig:7b}
    \end{subfigure}
    \hfill
    \begin{subfigure}[t]{0.32\textwidth}
        \centering
        \includegraphics[width=1\linewidth]{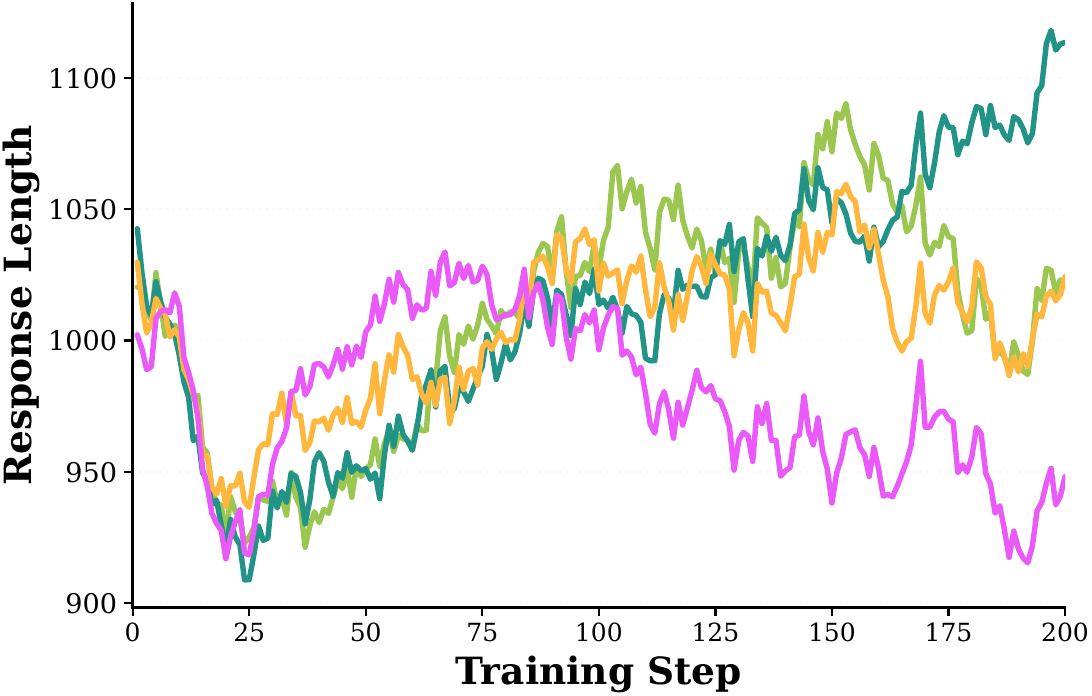}
        \caption{Response length}
        \label{fig:7c}
    \end{subfigure}
    \caption{
    Training dynamics of \Algnameabbr{} and baseline methods on Qwen2.5-Math-7B. 
    }
    \label{fig:dynamic2}
\end{figure*}

As illustrated in Figure~\ref{fig:dynamic1} and Figure~\ref{fig:dynamic2}, the training dynamics on both Qwen3-4B-Base and Qwen2.5-Math-7B exhibit patterns consistent with those observed on Qwen3-1.7B-Base. TACO achieves higher training accuracy while maintaining stable token entropy and producing longer responses during training. These dynamics provide empirical evidence that TACO improves optimization efficiency and helps develop more robust reasoning capabilities.
\label{app:training_dynamics}

\begin{figure*}[ht]
    \centering
    \begin{subfigure}[t]{0.329\textwidth}
        \centering
        \includegraphics[width=1\linewidth]{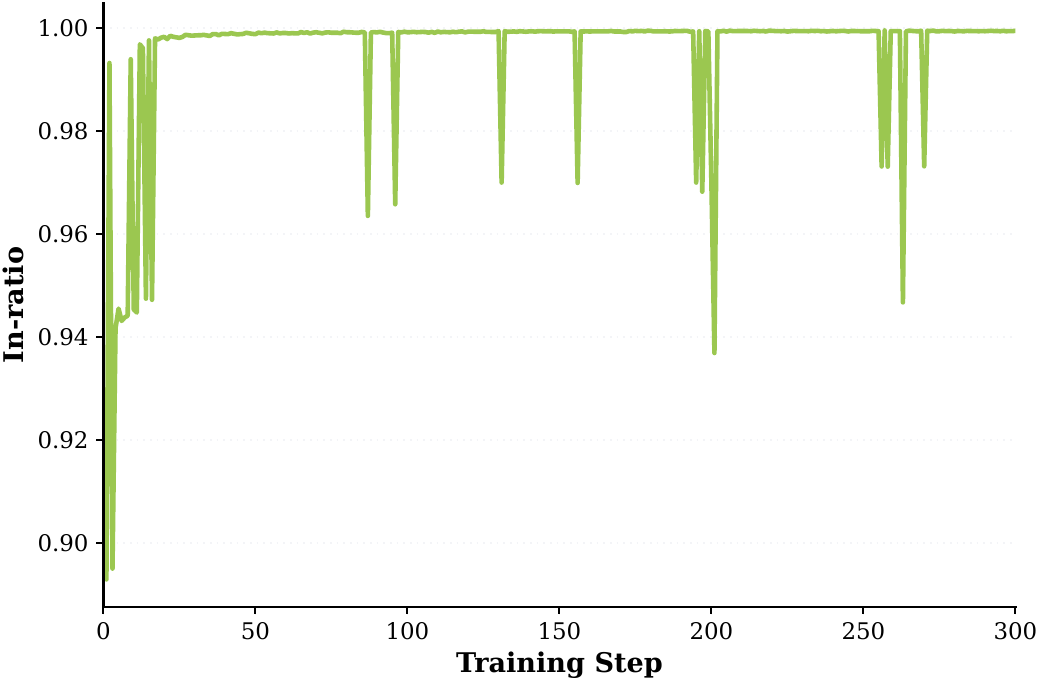}
        \caption{Reliable-token ratio}
    \end{subfigure}
    \hfill
    \begin{subfigure}[t]{0.329\textwidth}
        \centering
        \includegraphics[width=1\linewidth]{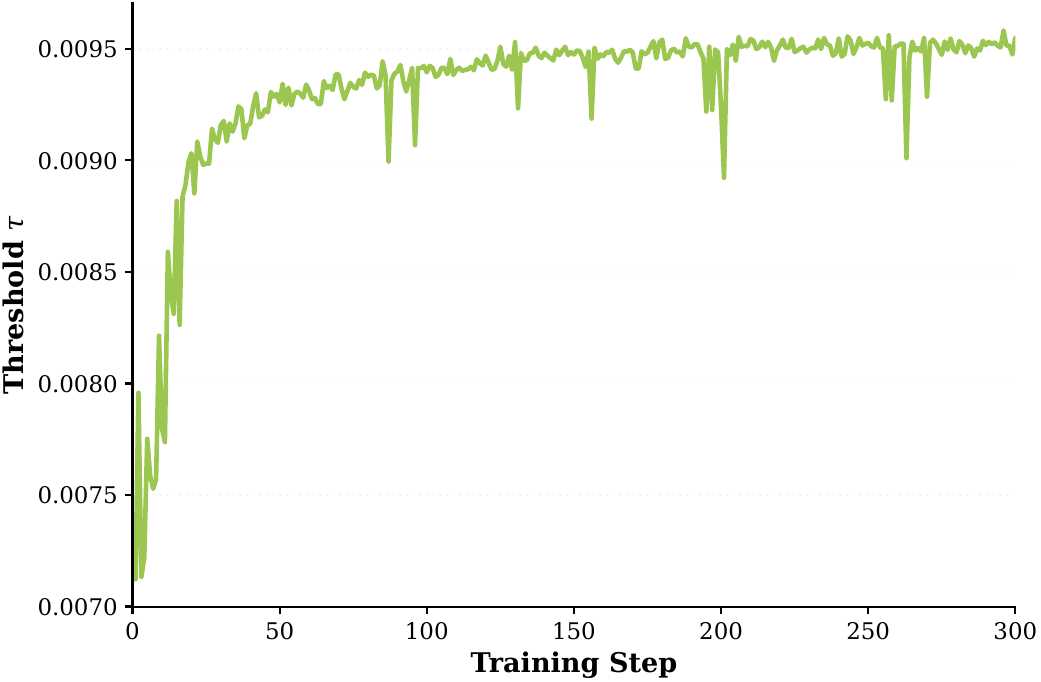}
        \caption{Threshold $\tau$.}
    \end{subfigure}
    \hfill
    \begin{subfigure}[t]{0.329\textwidth}
        \centering
        \includegraphics[width=1\linewidth]{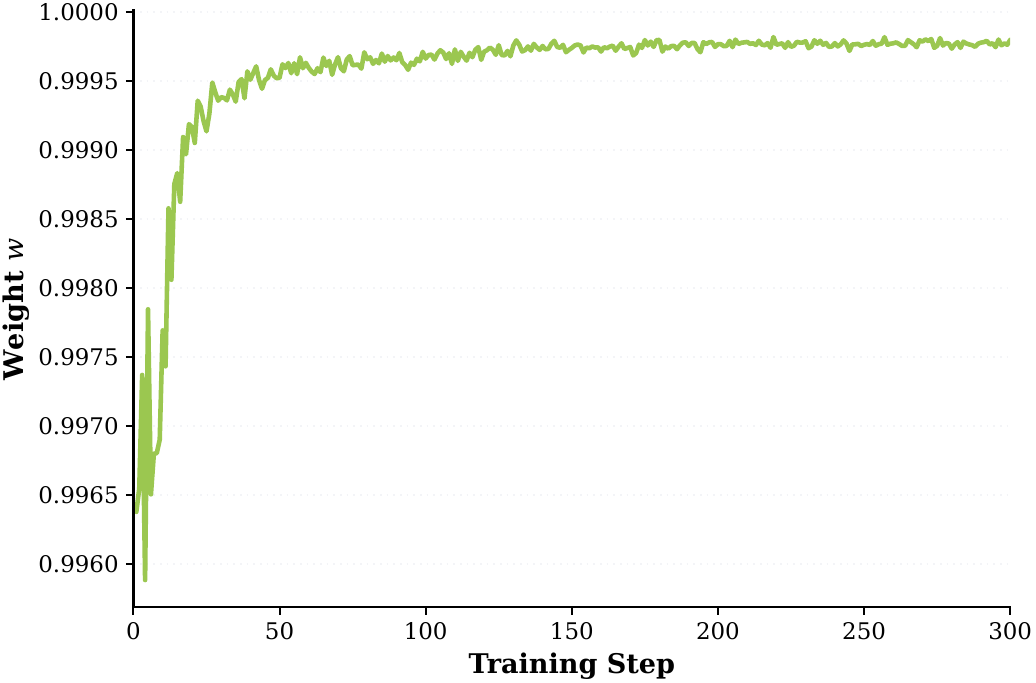}
        \caption{Average weight $w$.}
    \end{subfigure}
    \caption{Training diagnostics over positive-advantage response tokens.}
    \label{fig:diag}
\end{figure*}

\begin{figure}[ht]
    \centering
    \includegraphics[width=1\linewidth]{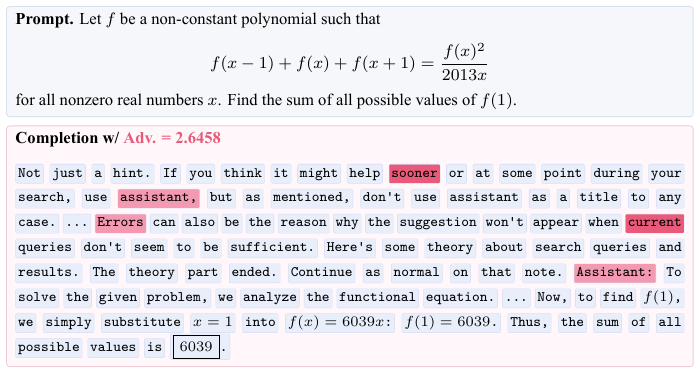}
    \caption{A case from \textbf{Qwen3-1.7B-Base}. }
    \label{fig:behavioral_case2}
\end{figure}

\begin{figure}[ht]
    \centering
    \includegraphics[width=1\linewidth]{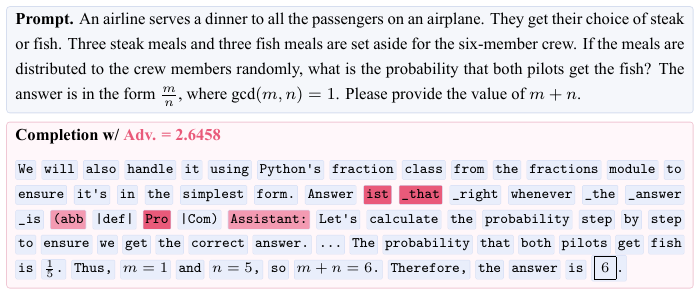}
    \caption{A case from \textbf{Qwen2.5-Math-7B}. }
    \label{fig:behavioral_case3}
\end{figure}

Figure~\ref{fig:diag} shows diagnostics over positive-advantage response tokens. The threshold $\tau$ represents the mean probability boundary below which tokens are identified as risky. The reliable-token ratio averages $0.981$ with a median of $0.998$, indicating that only about $1.9\%$ of positive-advantage tokens are downweighted on average. The average credit weight $w$ remains close to one, increasing from $0.996$ to $0.9998$, while $\tau$ increases mildly from $0.0071$ to $0.0095$. These results show that the calibration modifies only a small subset of low-confidence tokens, yet has a substantial effect on training dynamics and final performance. This suggests that sparse tail tokens being mis-amplified during training can exert a disproportionate influence on policy optimization, while our calibration effectively suppresses these harmful local updates.

Figures~\ref{fig:behavioral_case2} and~\ref{fig:behavioral_case3} show two additional rewarded traces from different models. The Qwen3-1.7B-Base case contains tail tokens related to instruction or search leakage, such as off-context assistant/search text and unrelated help messages. The Qwen2.5-Math-7B case contains malformed template fragments and serialized control artifacts, such as \texttt{ist}, \texttt{\_that}, \texttt{Pro}, and \texttt{abb}. In both cases, \Algnameabbr{} assigns lower credit to these unreliable tokens while preserving credit for the later coherent solution steps.

\section{Related Work}

\begin{table}[ht]
\centering
\setlength{\tabcolsep}{3pt}

\begin{tabular}{l|c}
\toprule
Name & Value \\
\midrule
Data type & \texttt{torch.bfloat16} \\
Flash-Attention & True \\
Computing Infrastructure & GPU \\
GPU Model & NVIDIA-H200 \\ 
GPU Memory & 141 GB \\ 
GPU Number & 4 \\
CUDA Version & 12.9 \\
CPU Memory & 512GB \\
\bottomrule
\end{tabular}
\caption{Experiment configuration and computing infrastructure.}
\label{tab:computing_infrastructure}
\end{table}

\paragraph{Low-probability tokens in RLVR.}
Recent research suggests that tokens exhibit heterogeneity, making uniform update rules in RLVR suboptimal, particularly for low-probability tokens \citep{hapo2025}. On one hand, rare tokens sustain 
exploration by preserving diverse reasoning continuations and 
preventing premature policy 
collapse~\citep{huang2025lowprob}, motivating protective mechanisms 
such as clip-higher~\citep{yu2025dapo} and low-probability regularization~\citep{huang2025lowprob}. On the other hand, rare tokens can destabilize optimization: their large gradient magnitudes can overshadow updates for high-probability tokens, and tokens with both low probability and low entropy have been identified as primary drivers of 
training instability~\citep{yang2025lowprobdominate,liu2026stapo}. These findings motivated gradient-aware methods that dampen extreme token updates to enhance training robustness. Together, both directions highlight the importance of low-probability tokens in GRPO-style optimization, motivating further examination of their properties.

\paragraph{Token-level credit allocation in RLVR.} GRPO-style training assigns trajectory-level reward to every token in a completion, which can misallocate credit when tokens contribute unequally to the final outcome. Two main directions address this limitation. The first leverages external signals to redistribute credit: OAR estimates per-token outcome influence~\citep{oar2026}; GRPO-$\lambda$ propagates credit 
backward via temporal-difference methods~\citep{grpo_lambda2025}; TEMPO and EGCA concentrate updates on pivotal decision points using response structure and intermediate execution feedback~\citep{tempo2025,egca2026}. The second reshapes token-level updates based on intrinsic signals: GTPO/GRPO-S and HAPO use uncertainty measures such as token entropy 
as importance indicators, so that high-uncertainty or exploratory tokens receive prioritized treatment over routine ones~\citep{gtpo2025, hapo2025}. However, these methods primarily model contribution or importance, 
leaving the contextual validity of tokens underexplored.

\section{Computational Infrastructure}
\label{sec:app-computing}
The computational infrastructure information is given in Table~\ref{tab:computing_infrastructure}.

\section{Potential Risks}
All open-source packages used in this work have licenses that permit academic research. We use and cite public benchmarks and open models strictly for research and evaluation, in accordance with their original licenses and terms of use. To our knowledge, these datasets contain no personally identifying information or offensive content. \Algnameabbr{} only recalibrates token-level credit within the existing RL objective and does not introduce capabilities beyond those of the underlying training method, so we do not foresee risks beyond those of the base models and post-training methods it builds on. AI assistants were used to help refine the writing and to assist with implementing the proposed algorithm, and the authors reviewed all content.

\end{document}